\documentclass[letterpaper]{article} 
\usepackage{aaai2026}  
\usepackage{times}  
\usepackage{helvet}  
\usepackage{courier}  
\usepackage[hyphens]{url}  
\usepackage{graphicx} 
\usepackage{natbib}  
\usepackage{caption} 
\usepackage{algorithm}
\usepackage{algorithmic}
\usepackage{booktabs}
\usepackage{multirow}
\usepackage{adjustbox}
\usepackage{graphicx}
\usepackage{kotex}
\usepackage{amsmath}
\usepackage{cleveref}
\usepackage{subcaption}
\usepackage{makecell} 
\usepackage{newfloat}
\usepackage{listings}
\DeclareCaptionStyle{ruled}{labelfont=normalfont,labelsep=colon,strut=off} 
\floatstyle{ruled}
\newfloat{listing}{tb}{lst}{}
\floatname{listing}{Listing}
\title{BD-Net: Has Depth-Wise Convolution Ever Been Applied in \\ Binary Neural Networks?}
\author {
    DoYoung Kim,
    Jin-Seop Lee,
    Noo-ri Kim,
    SungJoon Lee,
    Jee-Hyong Lee\thanks{Corresponding author}
}
\affiliations {
    Sungkyunkwan University, Suwon, South Korea\\
    \{kacel33, wlstjq0602, pd99j,  sjoon8379, john\}@skku.edu
}

\usepackage{bibentry}

\begin{document}

\maketitle

\begin{abstract}
Recent advances in model compression have highlighted the potential of low-bit precision techniques, with Binary Neural Networks (BNNs) attracting attention for their extreme efficiency. However, extreme quantization in BNNs limits representational capacity and destabilizes training, posing significant challenges for lightweight architectures with depth-wise convolutions.
To address this, we propose a 1.58-bit convolution to enhance expressiveness and a pre-BN residual connection to stabilize optimization by improving the Hessian condition number. These innovations enable, to the best of our knowledge, the first successful binarization of depth-wise convolutions in BNNs.
Our method achieves 33M OPs on ImageNet with MobileNet V1, establishing a new state-of-the-art in BNNs by outperforming prior methods with comparable OPs. Moreover, it consistently outperforms existing methods across various datasets, including CIFAR-10, CIFAR-100, STL-10, Tiny ImageNet, and Oxford Flowers 102, with accuracy improvements of up to 9.3 percentage points.
\end{abstract}

\begin{links}
    \link{Code}{https://github.com/kacel33/BD-Net}
\end{links}

\section{Introduction}
\label{sec:intro}

Deep Neural Networks (DNNs) have demonstrated significant performance improvements in various tasks, including image classification~\cite{he2016deep, rawat2017deep, Kim_2024_CVPR}, object detection~\cite{redmon2016you, zou2023object}, and semantic segmentation~\cite{long2015fully, minaee2021image}.
Despite this success, they require substantial parameters and computational cost for training and inference. 
Consequently, most DNNs require computational accelerators like high-end GPUs, making them unsuitable for edge devices in resource-constrained environments. 
Recently, there has been a growing demand for personalized on-device services using DNNs, highlighting the importance of developing lightweight models that can operate efficiently in these limited environments.
To achieve lightweight models, model compression methods have been proposed, such as knowledge distillation~\cite{hinton2015distilling,gou2021knowledge}, pruning~\cite{han2015learning, liang2021pruning}, compact networks~\cite{howard2019searching, zhang2018shufflenet, han2020ghostnet}, and low-bit precision~\cite{zhou2016dorefa, jacob2018quantization, rakka2022mixed, shin2023nipq}.

\begin{figure}[t]
    \includegraphics[width=0.48\textwidth,height=6cm]{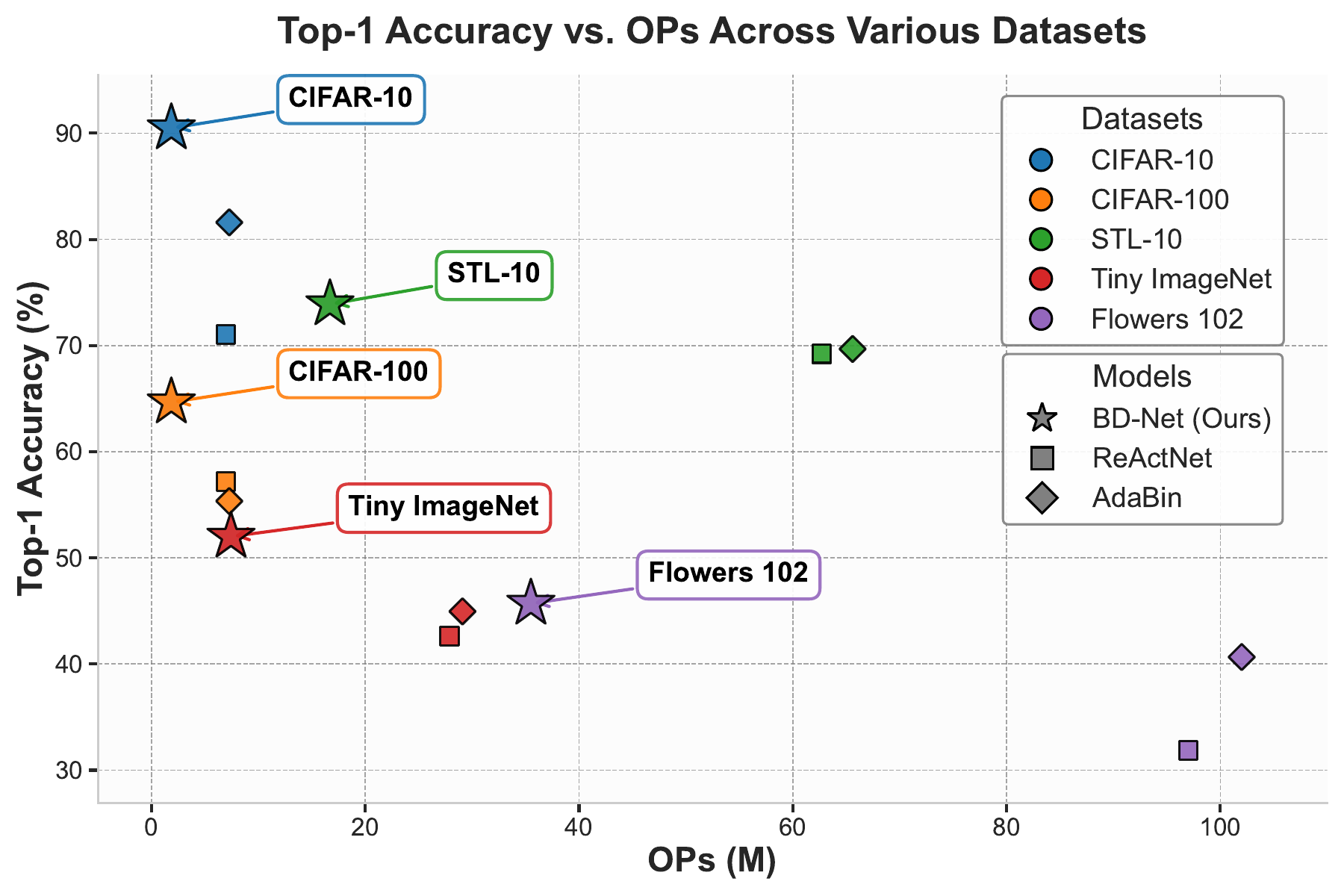} 
    \caption{Top-1 accuracy vs. operations (millions of OPs) across various datasets. Colors denote datasets: CIFAR-10 (blue), CIFAR-100 (orange), STL-10 (green), Tiny ImageNet (red), and Flowers-102 (purple). Marker shapes represent models: BD-Net (stars), ReActNet (squares), and AdaBin (diamonds). A blue star, for example, indicates BD-Net applied to CIFAR-10. BD-Net consistently achieves superior accuracy with significantly fewer operations across all datasets.}
    \label{fig:figure1}
\end{figure}

Recently, Binary Neural Networks (BNNs) have attracted significant attention because they can drastically reduce computational costs by quantizing both model weights and activations to 1-bit. 
Since they use 1-bit operations, traditional multiplication-accumulation operations can be replaced by XNOR-bitcount operations, making them highly effective in resource-constrained environments like low-power CPUs.
The key challenge in BNNs is minimizing performance degradation.
Since weights and activations are quantized to 1-bit precision, the network's representational capacity is significantly reduced, leading to performance degradation.
To address this issue, quantization approximations~\cite{rastegari2016xnor, lin2017towards, tu2022adabin} and adaptive thresholds~\cite{liu2020reactnet, lee2023insta} have been proposed to reduce quantization errors. 

However, these binarization approaches have proven effective only for DNNs with regular convolutions and have not been successfully applied to compact architectures such as the MobileNet series~\cite{howard2017mobilenets, sandler2018mobilenetv2, howard2019searching}.
Attempts to binarize depth-wise convolutions (DW convs) using existing approaches result in substantial performance degradation. Although DW convs are essential components for lightweight model architectures, current binarization techniques fail to effectively quantize these operations. To address these limitations of BNNs and enhance their applicability in creating more lightweight models, new specialized binarization methods specifically designed for depth-wise separable convolutions are needed.

The difficulty in effectively binarizing DW convs arises from a combination of two factors: the use of 1-bit precision and the structural limitation of DW convs. Usually, low-bit precision quantization techniques control the model performance by adjusting the number of bits for quantization~\cite{gluska2020exploring, wang2019haq}. However, in BNNs, which use only 1-bit precision, handling quantization errors becomes particularly challenging.
The structural characteristic of DW convs is the small number of parameters compared to regular convs. Due to this, it is challenging for quantization errors to be effectively mitigated~\cite{finkelstein2019fighting}. Additionally, regardless of quantization, due to the channel-wise independent convolution operation, DW convs often exhibit zero variances in batch normalization (BN) layers, leading to large gradient fluctuations~\cite{sheng2018quantization}. When this structural feature is combined with binarization, gradients may fluctuate more, resulting in a very rough loss landscape, and model training becomes very challenging.

We propose a method for effectively \textbf{binarizing DW conv operations in BNNs} while minimizing performance loss. To reduce quantization errors in BNNs, we introduce 1.58-bit conv operations to enhance representational capacity and a pre-BN residual connection to smooth the loss landscape, leading to more stable training.
These techniques enable the first successful binarization of DW convs in BNNs. 
As shown in Figure~\ref{fig:figure1}, our method significantly reduces OPs while simultaneously improving accuracy across multiple datasets, achieving accuracy gains of up to 9.3 percentage points (p.p.) on CIFAR-10, CIFAR-100, STL-10, Tiny ImageNet, and Oxford Flowers 102.
On ImageNet, it maintains competitive accuracy with approximately 3 times fewer OPs compared to regular conv-based BNNs.
This represents a significant reduction in computational cost compared to current state-of-the-art BNNs.

\section{Preliminaries}
\label{sec:preliminary}

In BNNs, the computational cost can be reduced by replacing traditional multiplication-accumulation operations with XNOR-bitcount operations. The total operations ($OP$) are determined by both binary operations ($BOP$) and floating point operations ($FLOP$), which are expressed as $OP = BOP/64 + FLOP$~\cite{bulat2019xnor, liu2018bi}. However, despite the significant compression ratio, conventional BNNs do not significantly reduce computational cost when binarizing 32-bit compact networks. This is mainly because DW conv has not been successfully binarized. In this section, we delve into the reasons why DW conv needs to be binarized, and explore the challenges that have prevented its successful binarization.

\begin{table}[t]
\centering
\renewcommand{\arraystretch}{1.2}
\begin{tabular}{ccc}
\toprule
\textbf{Type} & \textbf{Operation} &  \textbf{Count}   \\
\hline
\multirow{2}{*}{Full-precision} & 3$\times$3 regular conv. & 462M \\
 & 3$\times$3 depth-wise conv. & 3.61M  \\
\hline
\multirow{2}{*}{Binary} & 
3$\times$3 regular conv. & 7.23M  \\
& 3$\times$3 depth-wise conv. & 56K  \\
\bottomrule
\end{tabular}
\caption{Operation count comparison of convolutional operations for full-precision and binary networks. The counts are for a CNN layer where the resolution is 56$\times$56, and both the input and output channels are 128, respectively.}
\label{tab:concom}

\end{table}

\subsection{Why depth-wise convolution needs to be binarized}
The depth-wise separable structure introduced in the MobileNet series is a key technology for lightweight CNN models. In this structure, DW conv requires significantly less computation compared to regular convs. 

Table~\ref{tab:concom} provides a comparison of different conv operations along with their operation counts of a CNN layer.
Interestingly, binary regular convs still require more computation than 32-bit DW conv. 
While BNNs are more efficient than standard 32-bit models, they are not necessarily more efficient than compact models using 32-bit DW convs.

We can binarize DW conv similarly to regular convs. However, such a simple attempt frequently results in decreased performance. For this reason, early BNN research~\cite{liu2020reactnet, tu2022adabin, lee2023insta} replaced 3$\times$3 DW conv in MobileNet V1's backbone with 3$\times$3 regular convs. While this substitution improved performance, it did not significantly reduce computational cost. Theoretically, BNN approaches could reduce model size by up to 64 times, but ReActNet~\cite{liu2020reactnet} achieved only a 6.5-fold reduction, from 569M OPs in full-precision MobileNet V1 to 87M OPs. We performed experiments on ReActNet with DW conv and observed a substantial drop in performance, reducing Top-1 accuracy on ImageNet from 69.4\% to 60.7\%. Similar issues have also been observed in low-bit precision environments. In mixed precision settings, more bits are usually allocated in depth-wise separable convolutions~\cite{gluska2020exploring, wang2019haq, sheng2018quantization}. 
However, in BNNs, we cannot allocate more bits because we use only 1-bit precision.

Therefore, to build more efficient and lightweight models in BNNs, it is crucial to effectively binarize DW conv.

\subsection{Why depth-wise convolution hasn't been binarized}
Problems caused by quantization are closely related to the number of bits and the number of parameters \cite{finkelstein2019fighting}.
If a model is quantized with a small number of bits, the reduced representational capacity leads to lower model performance. Additionally, using a small number of quantization bits can cause the weights or activation values to fluctuate significantly near the quantization rounding boundaries during training. Such rapid changes lead to sudden changes in gradients, making the learning process unstable and slowing down convergence speed~\cite{nagel2022overcoming}.

However, when the number of parameters in a model is large, both the quantization errors from each parameter and the changes in weights and activations can be easily canceled out. As a result, the influence of quantization errors decreases, and the gradients become relatively stable, facilitating the model’s learning.
 
This is why previous studies have successfully binarized regular convs, but have not binarized DW conv, which has fewer learnable parameters.
Particularly in BNNs, the use of 1-bit quantization results in very high quantization errors, causing the gradients to change drastically. This makes it more challenging for models with fewer parameters to handle these effects properly.

Independent of quantization, DW conv has another problem due to its structural characteristics, which causes learning instability. In regular convs, each output channel is influenced by all input channels, but in DW conv, each channel receives only one input channel. Therefore, the outputs of a channel are likely to have small variances~\cite{sheng2018quantization}.
The output of a BN layer  can be described as follows:
\begin{equation}
y = \alpha \left( x-\mu \right) + \beta, \text{ where} \quad \alpha = \frac{\gamma}{\sqrt{\sigma^{2}+ \epsilon}}. 
\label{eq:1}
\end{equation}
Here, $y$ is the normalized output, $\gamma$ and $\beta$ are the scale and shift parameters, and $\alpha$ is the scaling factor. If $\sigma$ approaches zero, the scaling factor $\alpha$ can become abnormally large. 
It leads to large gradients, making the model training process potentially unstable~\cite{sheng2018quantization}.

\begin{figure}[t]
    \centering
    \begin{subfigure}[b]{0.23\textwidth}
        \centering
        \includegraphics[width=\textwidth]{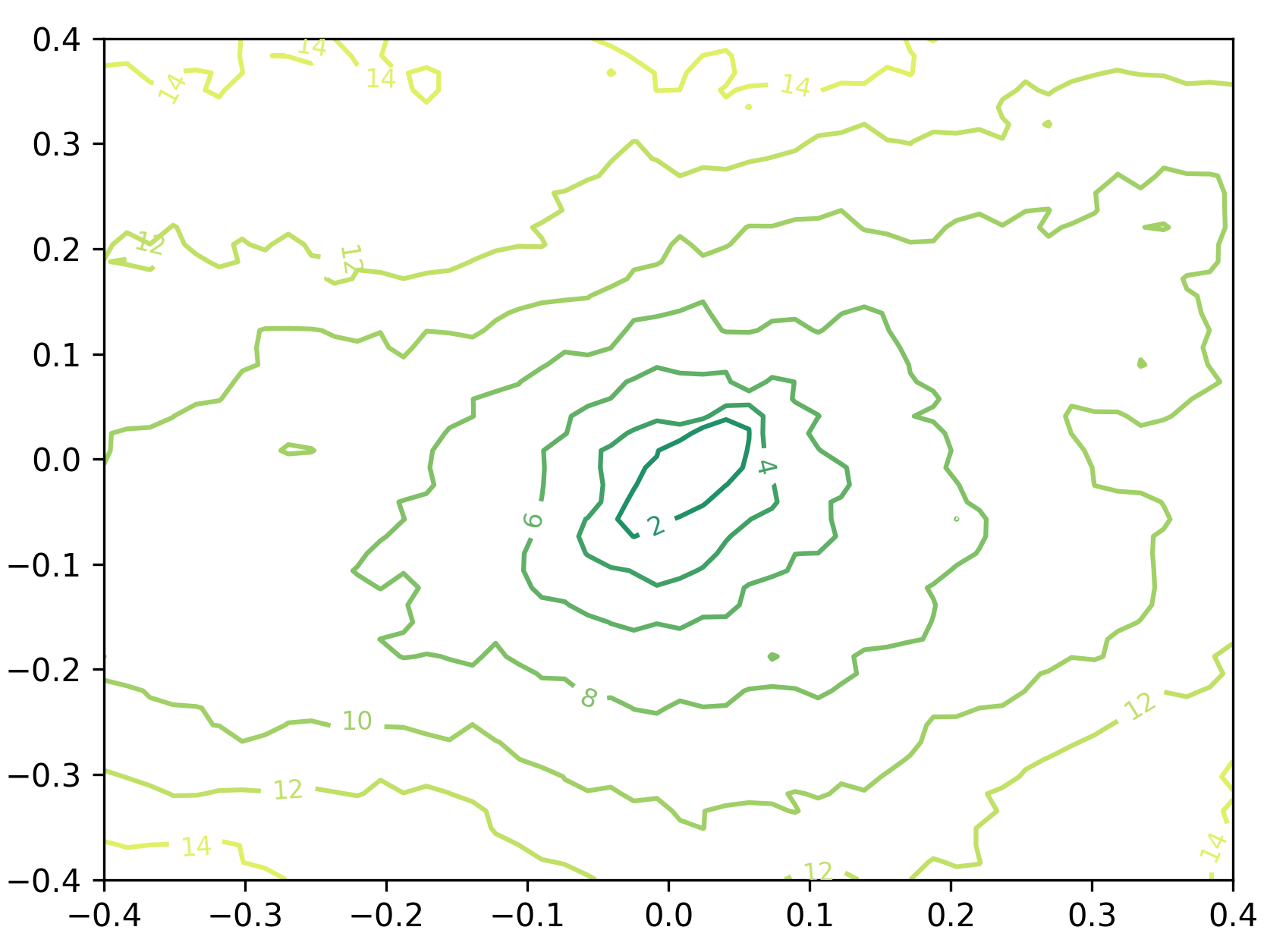}
        \caption{}
        \label{landscape_baseline}
    \end{subfigure}
    \begin{subfigure}[b]{0.23\textwidth}  
        \centering 
        \includegraphics[width=\textwidth]{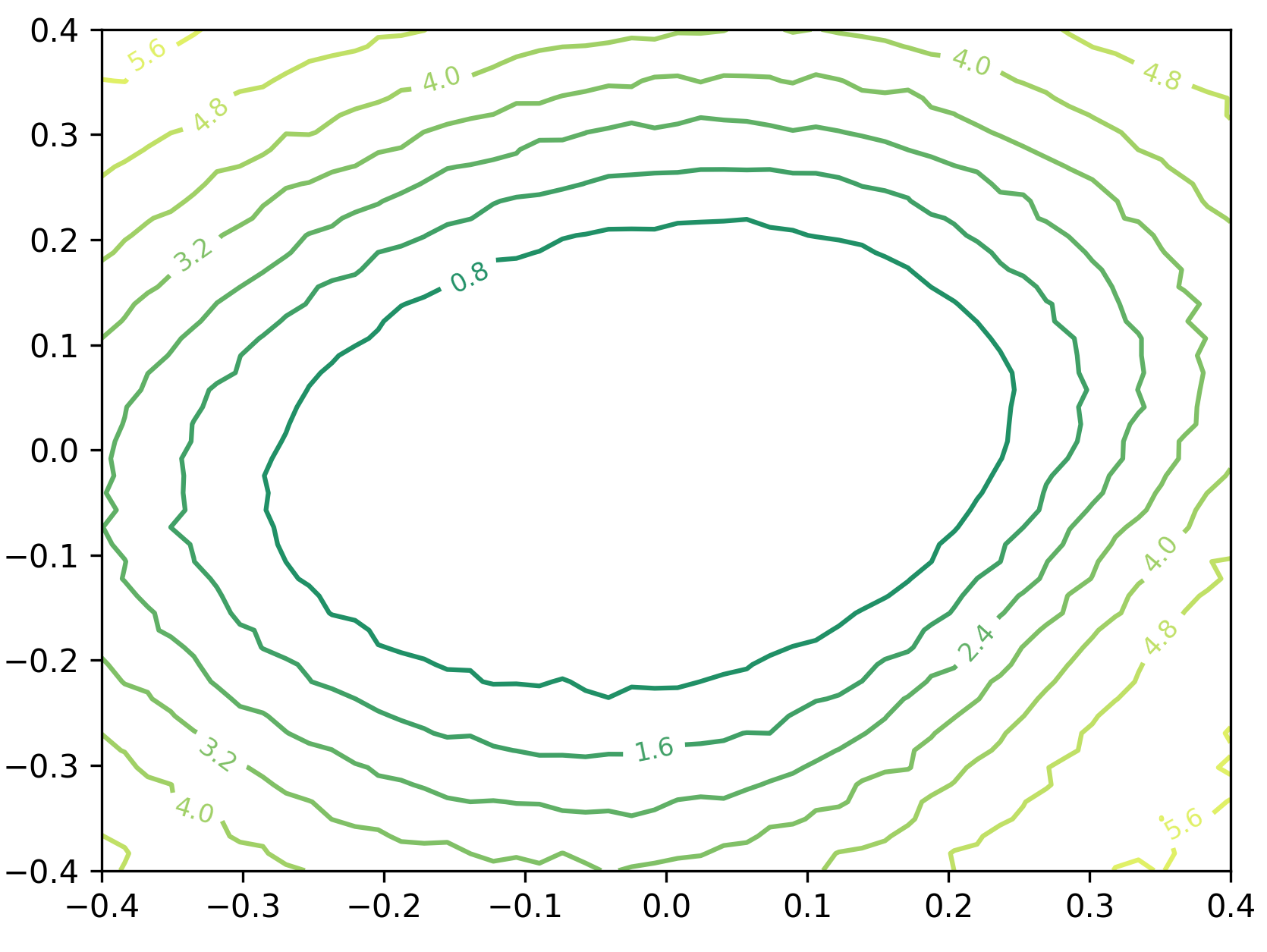}  
        \caption{}
        \label{landscape_32}
    \end{subfigure}
    \caption{Loss landscapes of (a) binarized MobileNet V1 and (b) full-precision MobileNet V1.}
    \label{fig:preliminary_landscape}
\end{figure}

The difficulties in binarizing DW conv can be summarized as follows:
\begin{itemize}
\item Due to fewer learning parameters, depth-wise convolution struggles to mitigate quantization errors, lowering the model's accuracy and increasing learning instability.
\item The 1-bit quantization, with only two values to represent, not only reduces the model performance, but also significantly increases learning instability due to extreme changes in weights during training.  
\item Depth-wise convolution is structurally more prone to learning instability. If it is binarized, the instability can be amplified.
\end{itemize}

\noindent
These three factors make binarized DW convs not only highly inaccurate, but also extremely difficult to train due to severe learning instability.

Figure~\ref{fig:preliminary_landscape} illustrates the loss landscapes of binarized MobileNet V1 and full-precision MobileNet V1. Despite the structural issues of DW conv, the full-precision version shows a relatively smooth landscape due to its full precision representation. However, the binarized MobileNet V1, affected by all the three factors, displays a very rough landscape. To successfully binarize DW conv with high performances, methods that can enhance expressiveness with only 1-bit quantization and smooth the loss landscape for training stability are required.

\begin{figure*}[t]
    \centering
    \begin{subfigure}[b]{0.38\textwidth}
        \centering
        \includegraphics[width=0.6\textwidth] 
        {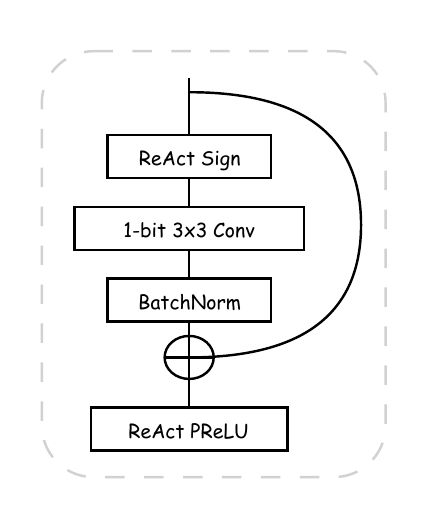} 
        \caption{ReActNet~\cite{liu2020reactnet}}
        \label{fig:first}
    \end{subfigure}
    \begin{subfigure}[b]{0.53\textwidth}
        \centering
        \includegraphics[width=0.6\textwidth]{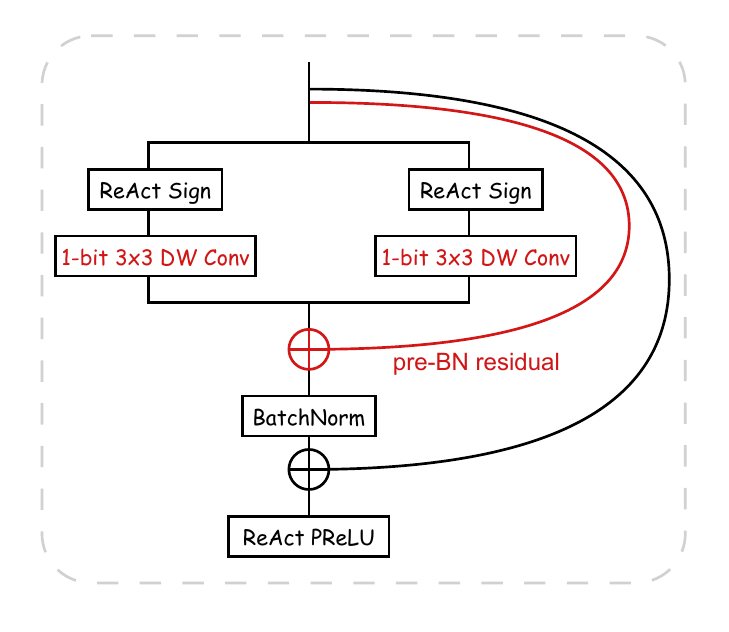} 
        \caption{Ours}
        \label{fig:second}
    \end{subfigure}
    \caption{Architecture comparison of binary depth-wise convolution layers.}
    \label{fig:twocol}
\end{figure*}

\section{Related Work}

Binary Neural Networks (BNNs) quantize both activations and weights to 1-bit, replacing traditional floating-point operations with faster bitwise operations. While this significantly accelerates computation, the reduction from 32-bit to 1-bit precision leads to information loss and resulting in performance degradation. To address these challenges and narrow the performance gap between BNNs and full-precision networks, methods such as quantization approximation~\cite{rastegari2016xnor, lin2017towards, tu2022adabin}, adaptive thresholding~\cite{liu2020reactnet, lee2023insta} and structural modification~\cite{liu2018bi, ma2024b} have been proposed.

In quantization approximation, XNOR-Net~\cite{rastegari2016xnor} introduces a real-valued scaling factor to improve approximation accuracy and ABC-Net~\cite{lin2017towards} simulates full-precision behavior using linear combinations of multiple binary convs. AdaBin~\cite{tu2022adabin} replaces fixed binary sets with adaptively obtained optimal binary sets. Additionally, adaptive thresholding techniques further reduce quantization errors. ReActNet~\cite{liu2020reactnet} introduces learnable parameters in the binary sign function with RSign and RPReLU, enabling dynamic adjustment during training. INSTA-BNN~\cite{lee2023insta} applies instance-aware thresholding using a statistical measure for more precise binarization. In terms of structural modifications, Bi-Real Net~\cite{liu2018bi} preserves information through residual connections, while A\&B BNN~\cite{ma2024b} eliminates 32-bit multiplication in BNNs with a removable mask layer and quantized RPReLU.

Despite these advances, binarizing DW convs has been particularly challenging due to structural characteristics that lead to training instability. This has resulted in performance drops and minimal FLOP savings compared to full-precision compact networks. To avoid the training instability of binarizing DW convs, MoBiNet~\cite{phan2020mobinet} and Binary MobileNet~\cite{phan2020binarizing} used group convs instead. However, they still have not fully leveraged the computational efficiency of DW convs.

\section{Methodology}

In this section, we propose methods to effectively binarize DW convs while maintaining high performance.
We first introduce the pre-BN residual structure, which stabilizes the training of binary DW convs, followed by the 1.58-bit technique, which enhances their representational capacity.

\subsection{Pre-BN residual connection on binary DW conv}
\label{pre-bn_method}
Binarizing DW convs is challenging due to structural constraints and extremely low bit widths.
While prior studies have proposed residual strategies to stabilize binary networks, these approaches have been limited to binary regular convs. To the best of our knowledge, binary DW convs have not been studied before, and their severe ill-conditioning problems remain unaddressed. In this work, we identify these challenges and introduce the pre-BN residual connection, a simple yet effective method designed specifically to alleviate the ill-conditioning of binary DW convs. As illustrated in Figure~\ref{fig:twocol}, this connection directly links the input of the binary DW conv to the input of the subsequent BN layer, smoothing the loss landscape and stabilizing optimization, all without introducing additional learnable parameters.

A rough loss landscape is closely associated with a high condition number of the Hessian matrix~\cite{marteau2019globally}.
The condition number of the Hessian matrix is defined as follows:
\begin{equation}
\kappa(H) = \|H\|\|H^{-1}\| = \frac{\lambda_1(H)}{\lambda_{n}(H)} 
\end{equation}
where $\lambda_1 (H)$ and $\lambda_n (H)$ denote the largest and smallest eigenvalues of $H$, respectively.
A higher condition number indicates greater sensitivity to perturbations and a more rugged loss landscape.

Let $J$ be the Jacobian of a naive binary DW conv with BN, and $J'$ be that of our BD-Net with the pre-BN residual. Their corresponding Hessians are denoted as $H$ and $H'$, respectively. The Jacobian $J$ can be expressed as:
\begin{equation}
J =  \alpha J^{dw}+ I
\end{equation}
where $\alpha$ is the scaling factor of the BN as defined in \Cref{eq:1}, and $J^{dw}$ is the Jacobian of the DW conv layer. The Jacobian of BD-Net is then formulated as:
\begin{equation}
J' =  \alpha J^{dw}+ (\alpha +1)I.
\end{equation}
Since the condition number of the Hessian approximately satisfies $\kappa(H) \approx {\kappa(J)}^2$, the condition numbers of $H$ and $H'$ can be approximated as:
\begin{equation}
\kappa (H) \approx  \Big\{ \frac{\lambda_1(J)}{\lambda_n(J)}\Big\}^2,
\end{equation}

\begin{equation}
\kappa (H') \approx  \Big\{ \frac{\lambda_1(J')}{\lambda_n(J')}\Big\}^2 = \Big\{\frac{\lambda_1(J)+\alpha}{\lambda_n(J) + \alpha} \Big\}^2.
\end{equation}
The condition number of $J'$ is computed as:
\begin{equation}
\begin{split}
\kappa (J') &= \frac{\lambda_{1}(J')}{\lambda_n(J')} = \frac{\lambda_1(J)+\alpha}{\lambda_n(J) + \alpha} \\
&= \frac{\lambda_1 (J) + \alpha}{\lambda_n (J) + \alpha} \cdot \frac{\lambda_n (J)}{\lambda_1 (J)} \cdot \frac{\lambda_1 (J)}{\lambda_n (J)} \\ 
&= \frac{\lambda_n (J) (\lambda_1 (J) + \alpha)}{\lambda_1 (J) (\lambda_n (J) + \alpha)} \cdot \frac{\lambda_1 (J)}{\lambda_n (J)} \\
&= \Big( 1 + \frac{\alpha}{\lambda_1 (J)} \Big) \cdot \Big( \frac{\lambda_n (J)}{\lambda_n (J) + \alpha} \Big) \cdot \frac{\lambda_1 (J)}{\lambda_n (J)}.
\end{split}
\label{condition_number_proof}
\end{equation}
In binary DW convs, the variance of the output is often near zero due to extreme quantization, which causes the BN scaling factor $\alpha$ to become significantly large. In such cases, since $\lambda_n(J) \ll \alpha$, we can approximate:
\begin{equation}
\kappa (J') \approx \Big( 1 + \frac{\alpha}{\lambda_1 (J)} \Big) \cdot \frac{\lambda_n(J)}{\alpha} \cdot \frac{\lambda_1 (J)}{\lambda_n (J)}.
\end{equation}
Consequently, the condition number of $J'$ becomes:
\begin{equation}
\kappa (J') \approx \Big( \frac{\lambda_n(J)}{\alpha} + \frac{\lambda_n(J)}{\lambda_1 (J)} \Big) \cdot \kappa (J).
\end{equation}
As $\alpha$ increases, the condition number of the DW conv layer with the pre-BN residual approaches $\frac{\lambda_n(J)}{\lambda_1(J)} \cdot \kappa(J)$, leading to a significant reduction. Therefore, the condition numbers satisfy:
\begin{equation}
\kappa(H') < \kappa(H).
\end{equation}
This demonstrates that the pre-BN residual systematically controls the condition number via $\alpha$, directly flattening the loss landscape and ensuring stable convergence during training. This structural solution effectively mitigates the limitations of binary DW convs with minimal computational overhead with no additional learnable parameters. The practical benefits of the pre-BN residual are further validated in the Experiments section through extensive experiments, which demonstrate notable improvements in both training stability and accuracy.

\subsection{1.58-bit convolution}
DW conv has been successfully implemented in full-precision, contributing to the lightweight design of CNN-based models. However, unlike 32-bit DW conv, binary DW conv suffers from critically limited filter diversity. For example, while binary regular conv can generate $2^{9 \times C_{in}}$ filter combinations, binary DW conv is restricted to only $2^9$ combinations of $\{-1, 1\}$ filters. This limitation significantly reduces the network's ability to capture complex features, leading to poor representational capacity.

To address this limitation, we propose a dual binary conv structure that connects two binary convs in parallel and sums their outputs. Unlike prior work such as ABC-Net~\cite{lin2017towards}, which improves performance through linear combinations of multiple binary convs at the cost of excessive computational overhead (up to 25 times more convs), our method requires only a small increase in computation while significantly enhancing representational power.

In our approach, each binary conv has its own rounding boundary ($\alpha$) and output value ($\pm \beta$), and the final output is determined by combining the results of two binary convs. When $N$ binary convs are used, the number of output combinations becomes $N+1$. If we define the effective bit precision as $M = \log_2(N+1)$, the representational effect of the output grows logarithmically with respect to $N$. For example, when $N = 2$, we obtain $M = \log_2(3) \approx 1.58$, which inspires the term ``1.58-bit conv."

\begin{table*}[htbp]
\scriptsize
\centering
\renewcommand{\arraystretch}{0.75}
\begin{adjustbox}{width=\textwidth}
\begin{tabular}{ccccccccccc}
\toprule
\multirow{2}{*}{\textbf{Network}} & \multicolumn{2}{c}{\textbf{CIFAR-10}} & \multicolumn{2}{c}{\textbf{CIFAR-100}}& \multicolumn{2}{c}{\textbf{STL-10}} & \multicolumn{2}{c} {\textbf{Tiny ImageNet}} & \multicolumn{2}{c}{\textbf{Flowers 102}} \\
\cmidrule(lr){2-3} \cmidrule(lr){4-5} \cmidrule(lr){6-7} \cmidrule(lr){8-9} \cmidrule(lr){10-11}
 & \textbf{Acc} & \textbf{OPs} & \textbf{Acc} & \textbf{OPs} & \textbf{Acc} & \textbf{OPs} & \textbf{Acc} & \textbf{OPs} & \textbf{Acc} & \textbf{OPs} \\
\midrule
Full-Precision & 92.26 & 42.6M & 69.62 & 42.6M & 65.97 & 383M & 58.04 & 170M & 50.31 & 569M  \\
\midrule
ReActNet & 71.04  & 6.97M & 57.19  & 6.97M & 69.23 & 62.7M & 42.62 & 27.9M & 31.88 & 97M \\

\textbf{BD-Net-A (ReActNet)} & 78.37 & 1.56M & 50.11 & 1.56M & 65.60 & 14.0M & 38.98 & 6.23M & 36.46 & 32.5M \\
\textbf{BD-Net-B (ReActNet)} & \textbf{89.93} & 1.83M & \textbf{63.83} & 1.83M & \textbf{77.03} & 16.4M & \textbf{52.06} & 7.31M & \textbf{45.18} & 34.6M \\
\midrule
AdaBin & 81.60 & 7.28M & 55.35 & 7.28M & 69.68 & 65.6M & 44.96 & 29.1M & 40.66 & 102M \\
\textbf{BD-Net-A (AdaBin)} & 84.87 & 1.59M & 58.28 & 1.59M & 65.19 & 14.3M & 44.10 & 6.37M & 41.16 & 33M \\
\textbf{BD-Net-B (AdaBin)} & \textbf{90.48} & 1.86M & \textbf{64.66} & 1.86M & \textbf{73.90} & 16.7M & \textbf{52.01} & 7.44M & \textbf{45.70} & 35.5M \\
\bottomrule
\end{tabular}
\end{adjustbox}
\caption{Experimental Results for Different Networks on Various Datasets}
\label{tab:small-dataset-results}
\end{table*}


\begin{table*}[t]
    \centering
    \renewcommand{\arraystretch}{1.0}
    \begin{adjustbox}{max width=\textwidth}
    \begin{tabular}{lcccccccccc}
        \toprule
        Method  & Backbone
        &
        Conv type &\makecell{BOPs \\ ($\times$10\textsuperscript{9})}& \makecell{FLOPs \\ ($\times$10\textsuperscript{8})} & \makecell{OPs \\ ($\times$10\textsuperscript{8})} & \makecell{Top-1 \\Acc. (\%)} & \makecell{Top-5 \\Acc. (\%)}\\[-0.2em]
        \midrule
        BNN  & ResNet & regular & 1.70 & 1.20 & 1.47 & 42.2 & 67.1   \\
        XNOR-Net  & ResNet &  regular & 1.68 & 1.40 & 1.66 & 51.2 & 73.2  \\
        Bi-real Net 18 & ResNet &  regular & 1.68 & 1.41 & 1.67 & 56.4 & 79.5  \\
        XNOR-Net++  & ResNet &  regular & 1.68 & 1.42 &1.68& 57.1 & 79.9  \\
        IR-NET & ResNet &  regular & 1.68 & 1.41 &1.67 & 58.1 & 80.0   \\
        Real-to-Binary & ResNet & regular & 1.68 & 1.40 & 1.66 & 65.4 & 86.2   \\
        ReCU  & ResNet &  regular & 1.68 & 1.40 &1.67 & 66.4 & 86.5   \\
        INSTA-BNN  & ResNet & regular & 1.68 & 1.43 &1.70 & 67.6 & 87.5 \\[-0.2em]
        \midrule
        ReActNet  & MobileNet V1 & regular & 4.82 & 0.12 & 0.87 & 69.4 & -   \\
        Adabin  & MobileNet V1 &  regular & 4.82& 0.21 & 0.96 & 70.4 & -  \\
        INSTA-BNN  & MobileNet V1 &  regular & 4.82 & 0.20 & 0.95 & 71.7 & 90.3 \\[-0.2em]
        \midrule
        MoBiNet (K=4) & MobileNet V1 &  group & - & - & 0.52 & 54.4 & 77.5 \\
        MoBiNet (K=3) & MobileNet V1 &  group & - & - &0.49& 53.5 & 76.5   \\
        Binary MobileNet & MobileNet V1 &  group & - & - &0.33& 51.1 & 74.2  \\
        *ReActNet 0.5$\times$ & MobileNet V1 &  regular & 1.22 & 0.12 & 0.31 & 60.7 & 82.6  \\[-0.2em]
        \midrule
        BD-Net-A (ReActNet) & MobileNet V1 &  depth-wise & 1.09 & 0.13 & 0.30 & 63.9 & 84.8  \\
        BD-Net-B (ReActNet) & MobileNet V1 &  depth-wise & 1.08 & 0.16 & 0.33 & 65.3 & 85.8 \\
        BD-Net-B(x1.5) (ReActNet) & MobileNet V1 &  depth-wise & 2.50 & 0.26 & 0.65 & 69.4 & 88.8 \\[-0.2em]    
        \bottomrule
    \end{tabular}
    \end{adjustbox}
    \caption{Comparison of various methods for different metrics. The reproduced result is marked with *.}
    \vspace{-0.1cm}
    \label{tab:sota-imagenet}
\end{table*}
\begin{figure}[t]
    \centering
    \begin{subfigure}{.16\textwidth}
        \centering
        \includegraphics[width=.95\linewidth]{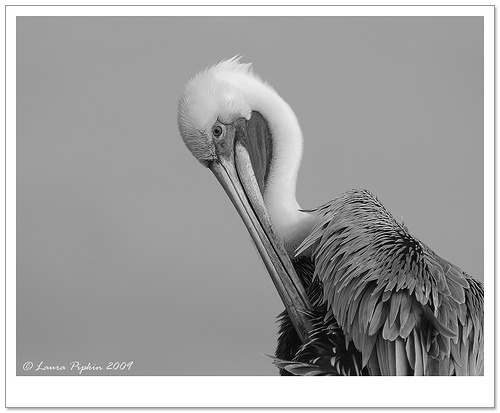}
        \caption{}
        \label{fig:sub1}
    \end{subfigure}%
    \begin{subfigure}{.16\textwidth}
        \centering
        \includegraphics[width=.95\linewidth]{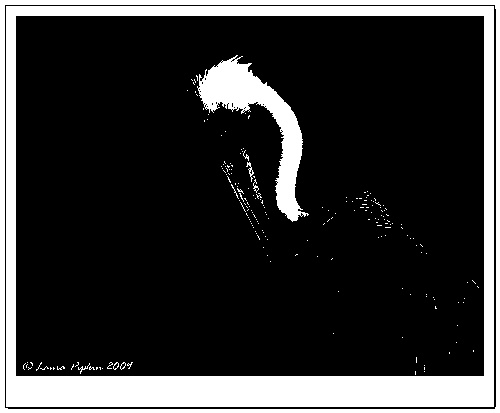}
        \caption{}
        \label{fig:sub2}
    \end{subfigure}%
    \begin{subfigure}{.16\textwidth}
        \centering
        \includegraphics[width=.95\linewidth]{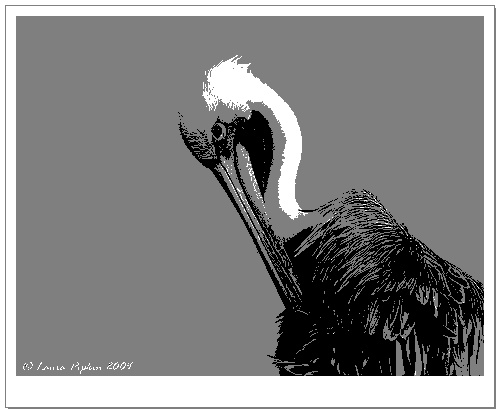}
        \caption{}
        \label{fig:sub3}
    \end{subfigure}
    \caption{Image visualization on ImageNet validation dataset. (a) Converted to grayscale, (b) Binarized using Otsu's algorithm, (c) Displayed using an appropriate threshold to represent in 1.58-bit.}
    \label{fig:1.58-bit-img}
\end{figure}
\begin{equation}
f(x)= 
\begin{cases}
-\beta_1 -\beta_2 & \text{if } x < \alpha_1, \\
\beta_1 - \beta_2 & \text{if } \alpha_1 \leq x < \alpha_2, \\
\beta_1 + \beta_2 & \text{if } \alpha_2 \leq x.
\end{cases}
\end{equation}

It is important to clarify that 1.58-bit equivalent does not refer to the actual bitwidth of weights or activations used in the network. Instead, it reflects the increased representational capacity achieved by combining multiple binary convs while maintaining bitwise operations. In this sense, the term expresses the enhanced expressive power of the dual binary conv rather than the physical storage or computational bit precision.

While $N = 3$ would yield an effect similar to 2-bit precision, using more than two binary convs introduces greater computational cost without proportional benefits. Therefore, we adopt the dual binary conv ($N = 2$) as an efficient trade-off between computation and representational capacity.

This method preserves the inherent efficiency of bitwise operations in BNNs while improving expressiveness. As illustrated in Figure~\ref{fig:1.58-bit-img}, conventional BNNs with a single rounding boundary may suffer from severe distortions in the shape and color representation of objects due to aggressive binarization. In contrast, the 1.58-bit equivalent conv alleviates such distortions to some extent, providing improved representational capacity while maintaining the efficiency of binary operations.

\section{Experiments}
\label{sec:experiments}
We primarily aim to explore the feasibility and potential benefits of incorporating DW convs into BNNs. To achieve this, we conduct experiments on CIFAR-10, CIFAR-100~\cite{krizhevsky2009learning}, STL-10~\cite{coates2011analysis}, Tiny ImageNet~\cite{le2015tiny}, Oxford Flowers 102~\cite{nilsback2008automated}, and ILSVRC12 ImageNet \cite{russakovsky2015imagenet}. 

To demonstrate the broader applicability of our approach, we implement our components on both ReActNet and AdaBin architectures, confirming the effectiveness of our method across different network designs.
BD-Net-A (ReActNet) and BD-Net-A (AdaBin) modify ReActNet and AdaBin, respectively, by replacing
the 3$\times$3 regular convs with 3$\times$3 DW convs and incorporating Pre-BN residual connections and 1.58-bit convs. Additionally, we introduce BD-Net-B (ReActNet) and BD-Net-B (AdaBin), variants of our method that replaces 1-bit 3$\times$3 DW convs in downsampling layers with real-valued 3$\times$3 DW convs in ReActNet and AdaBin, respectively.
For a fair comparison, the first and last layers are kept in full precision in all cases.

\subsection{Experimental Results}
To demonstrate the effectiveness and applicability of our proposed method, we conduct experiments on small- and medium-scale datasets such as CIFAR-10, CIFAR-100, STL-10, Tiny ImageNet, and Oxford Flowers 102, as well as on a large-scale dataset, ImageNet. 
The detailed implementation settings are provided in Appendix A.

\vspace{0.6em}
\noindent
\textbf{Small- and Medium-scale Datasets.}
Due to the challenges of binarizing DW convs, previous BNNs primarily relied on ResNet-18 or MobileNet V1 with regular convs replacing depth-wise operations.  In contrast, our method successfully binarizes DW convs while maintaining competitive performance. 

\Cref{tab:small-dataset-results} shows the experimental results. 
BD-Net-A (ReActNet) reduces computational cost by approximately 3$\times$ to 4.5$\times$ compared to ReActNet while maintaining competitive performance. Similarly, BD-Net-A (AdaBin) improves accuracy by +3.27 p.p., +2.93 p.p., and +0.50 p.p. on CIFAR-10, CIFAR-100, and Oxford Flowers 102, respectively, while reducing OPs by 26\%, 22\%, and 34\% on CIFAR datasets, Tiny ImageNet, and Oxford Flowers 102. 
Building on this, BD-Net-B (ReActNet) slightly increases OPs over BD-Net-A but consistently outperforms ReActNet across all datasets.
Likewise, BD-Net-B (AdaBin) further improves accuracy over BD-Net-A with only a marginal increase in OPs. Specifically, BD-Net-B achieves accuracy gains of +4.22 p.p. to +9.31 p.p. on CIFAR-10, CIFAR-100, STL-10, Tiny ImageNet, and Oxford Flowers 102, while reducing OPs by a factor of 2.68 to 8.47.

\vspace{0.4em}
\noindent\textbf{Large-scale Dataset.}
\Cref{tab:sota-imagenet} compares the performance of various methods on ImageNet, including ResNet-based methods, MobileNet V1-based methods with regular convs, and group conv-based methods.

ResNet-based methods achieve high accuracy but suffer from heavy computational costs. MobileNet V1-based methods, such as ReActNet, offer a better trade-off but still require more operations than group conv methods, which improve efficiency at the cost of reduced accuracy.

Our approach, BD-Net, addresses the limitations of previous methods by successfully incorporating DW convs into Binary Neural Networks. BD-Net-B achieves a Top-1 accuracy of 65.3\% with only 33M OPs on ImageNet, reducing the number of operations by approximately three times compared to MobileNet V1 with regular convs. 

To provide a fair comparison with approaches requiring similar computational resources, we evaluated ReActNet with half the number of channels. ReActNet\,0.5$\times$ achieves a top-1 accuracy of 60.7\% with 30M operations. In comparison, our method delivers over 4.6 p.p. higher performance with similar computational demands. This demonstrates that BD-Net not only reduces the required operations but also achieves superior performance through a more efficient architectural design.

\begin{table*}[t]
\scriptsize
\centering
\begin{adjustbox}{width=\textwidth}
\begin{tabular}{ccccccccccc}
\toprule
\multirow{2}{*}{\textbf{ShuffleNet V1}}  & \multicolumn{2}{c}{\textbf{CIFAR-10}} & \multicolumn{2}{c}{\textbf{CIFAR-100}}& \multicolumn{2}{c}{\textbf{STL-10}} & \multicolumn{2}{c} {\textbf{Tiny ImageNet}} & \multicolumn{2}{c}{\textbf{Flowers 102}} \\
\cmidrule(lr){2-3} \cmidrule(lr){4-5} \cmidrule(lr){6-7} \cmidrule(lr){8-9} \cmidrule(lr){10-11}
 & \textbf{Acc} & \textbf{OPs} & \textbf{Acc} & \textbf{OPs} & \textbf{Acc} & \textbf{OPs} & \textbf{Acc} & \textbf{OPs} & \textbf{Acc} & \textbf{OPs} \\

\midrule
Full-Precision & 91.16 & 9.70M & 67.16 & 9.70M & 67.12 & 23.3M & 49.88 & 10.3M & 40.97 & 126M \\
\midrule
ReActNet & 17.33 & 1.32M & 37.87 & 1.32M & 59.41 & 2.97M & 24.42 & 1.32M & 37.99 & 16.2M \\
ReActNet w/ br & 79.80 & 1.32M & 51.77 & 1.32M & 60.42 & 2.97M & \textbf{40.10} & 1.32M & \textbf{45.02} & 16.2M \\
\textbf{BD-Net-B (ReAct w/ br)} & \textbf{81.57} & 0.87M  & \textbf{59.03} & 0.87M & \textbf{64.31} & 1.96M & 35.83 & 0.87M & 44.10 & 10.7M \\
\bottomrule
\end{tabular}
\end{adjustbox}
\caption{Experimental results for Shufflenet V1 on various datasets. `br' indicates the use of broadcast residual connections, where previous layer channels are broadcast to match the next layer's channel dimensions for residual connections across layers with different channel counts.}
\label{tab:shufflenet_results}
\end{table*}

\begin{table*}[t]
\scriptsize
\centering
\begin{adjustbox}{width=\textwidth}
\begin{tabular}{ccccccccccc}
\toprule
\multirow{2}{*}{\textbf{MobileNet V3}}  & \multicolumn{2}{c}{\textbf{CIFAR-10}} & \multicolumn{2}{c}{\textbf{CIFAR-100}}& \multicolumn{2}{c}{\textbf{STL-10}} & \multicolumn{2}{c} {\textbf{Tiny ImageNet}} & \multicolumn{2}{c}{\textbf{Flowers 102}} \\
\cmidrule(lr){2-3} \cmidrule(lr){4-5} \cmidrule(lr){6-7} \cmidrule(lr){8-9} \cmidrule(lr){10-11}
 & \textbf{Acc} & \textbf{OPs} & \textbf{Acc} & \textbf{OPs} & \textbf{Acc} & \textbf{OPs} & \textbf{Acc} & \textbf{OPs} & \textbf{Acc} & \textbf{OPs} \\

\midrule
Full-Precision & 91.34 & 25.9M & 72.29 & 25.9M & 76.47
& 58.2M & 59.76 & 25.9M & 52.21 & 83.3M \\
\midrule
ReActNet & 19.08 & 9.80M & 14.96 & 9.80M & 24.08 & 6.26M & FAIL & 9.80M & 8.01 & 34.1M \\
ReActNet w/ br & \textbf{86.21} & 9.80M & \textbf{61.20} & 9.80M & 67.24 & 6.26M & 32.44 & 9.80M & 49.64 & 34.1M \\
\textbf{BD-Net-B (ReActNet w/ br)} & 85.92 & 1.14M & 58.60 & 1.14M & \textbf{69.22} & 1.47M & \textbf{42.41} & 1.14M & \textbf{51.33} & 7.99M \\
\bottomrule
\end{tabular}
\end{adjustbox}
\caption{Experimental results for MobileNet V3 on various datasets. `br' indicates the use of broadcast residual connections, where previous layer channels are broadcast to match the next layer's channel dimensions for residual connections across layers with different channel counts.}
\label{tab:mobilenetv3_results}
\end{table*}

\vspace{0.2em}
\noindent\textbf{Extension to Other Efficient Architectures.}

Due to the extreme reduction in representation capacity caused by binarization, existing BNNs have been applied only to ResNet and MobileNet V1. These networks have minimal variation in the number of channels across layers, enabling the existing layer-wise residual connections to be applied effectively.
However, in ShuffleNet V1 and MobileNet V3, the channel dimensions vary continuously across layers, preventing the application of existing layer-wise residual connections and resulting in unstable training. To address this, we propose broadcast residual connections. This method constructs residual connections by replicating the channels of the previous layer to match the channel dimensions of the next layer, enabling extension to ShuffleNet V1 and MobileNet V3.

Tables~\ref{tab:shufflenet_results} and~\ref{tab:mobilenetv3_results} show the performance measurements of ReActNet and BD-Net-B on ShuffleNet V1 and MobileNet V3. Simply applying ReActNet resulted in highly unstable training and significant accuracy degradation. However, with the addition of broadcast residual connections, accuracy improvements ranged from 1.01 p.p. to 62.47 p.p. on ShuffleNet V1 and from 32.44 p.p. to 67.13 p.p. on MobileNet V3 across datasets. This is because layer-wise residual connections increase representation capacity, enabling the model to learn more diverse features.
Nevertheless, applying ReActNet to ShuffleNet V1 and MobileNet V3 has limitations, as it reduces the OP count by only a factor of 2.4$\times$ to 9.3$\times$ by replacing 32-bit depth-wise convolutions with binary regular convolutions. In contrast, BD-Net-B achieves accuracy improvements ranging from 4.89 p.p. to 64.24 p.p. on ShuffleNet V1 and from 42.21 p.p. to 66.84 p.p. on MobileNet V3 while requiring significantly fewer operations than ReActNet.
Further improvements are required to effectively extend these methods beyond MobileNet V1 to the entire ShuffleNet and MobileNet series.

\begin{table}[t]
\centering
\begin{tabular}{l c}
\toprule
\textbf{Method} & \textbf{Accuracy (\%)} \\
\midrule
Baseline & 54.94 \\
+ Pre-BN & 56.93 \\
+ 1.58-bit conv & 56.18 \\
\textbf{Ours} & \textbf{58.28} \\
\bottomrule
\end{tabular}
\caption{Ablation study of our proposed methods on MobileNet V1 with CIFAR-100 dataset.}
\label{tab:ablation}
\end{table}
\subsection{Ablation Study}
To validate the effectiveness of each component in our proposed BD-Net architecture, we conduct an ablation study using MobileNet V1 on the CIFAR-100 dataset. As shown in Table~\ref{tab:ablation}, we use a naive binary DW conv as the baseline, achieving 54.94\% accuracy.
Introducing the Pre-BN residual connection improves accuracy to 56.93\% yielding a +1.99 p.p. gain. This result supports our theoretical analysis, where the Pre-BN residual structure stabilizes optimization by improving the Hessian condition number. Replacing the naive binary DW conv with the proposed 1.58-bit conv increases accuracy to 56.18\%, a +1.24 p.p. improvement over the baseline. This demonstrates that slightly increasing bit-width enhances the representational capacity of DW convs while maintaining a low computational cost.

Finally, combining both techniques in BD-Net achieves the highest accuracy of 58.28\%, with a +3.34 p.p. improvement over the baseline. These results demonstrate the complementary effects of the Pre-BN residual connection, which stabilizes optimization, and the 1.58-bit conv, which enhances feature representation. 
More extensive ablation studies on multiple convs are provided in Appendix B.

\subsection{Analysis}
\begin{figure}[t]
    \includegraphics[width=0.45\textwidth]{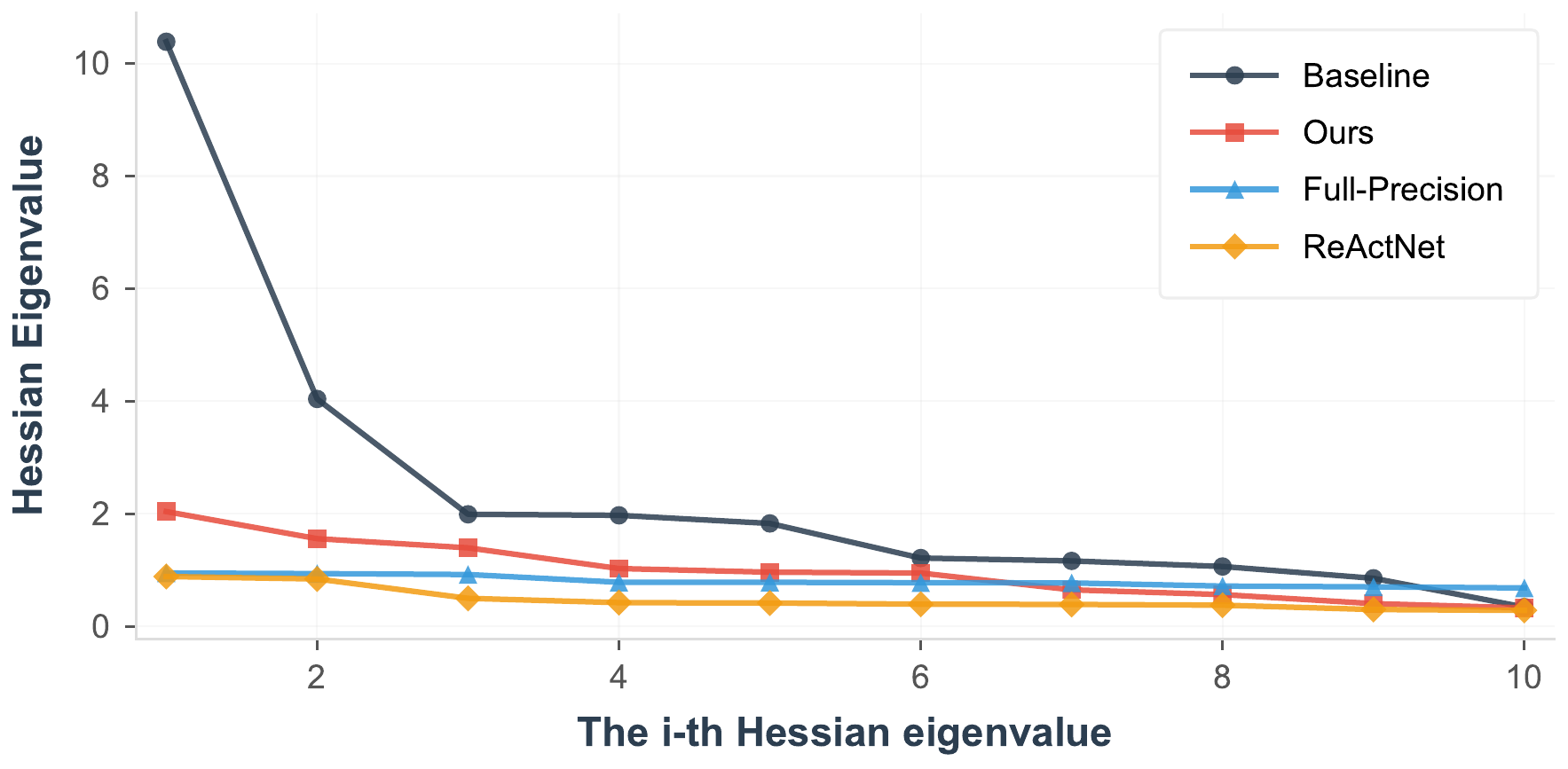} 
    \caption{Graph of the top-10 hessian eigenvalues for several networks.}
    \label{fig:hessian_graph}
\end{figure}

\noindent\textbf{Hessian Eigenvalue.}
Figure~\ref{fig:hessian_graph} illustrates the top-10 Hessian eigenvalues of the MobileNet V1 backbone across various methods: Baseline (simply binarized MobileNet V1), ReActNet, Full-precision, and our proposed approach. The graph shows that the baseline method, in comparison to the full-precision network, presents a notably higher maximum eigenvalue, leading to significant training instability. Our proposed method effectively reduces the maximum Hessian eigenvalue, thereby lowering the condition number of the Hessian and improving training stability. 
Moreover, our BD-Net exhibits an eigenvalue distribution similar to ReActNet with more parameters than ours, and to MobileNet in full precision. This indicates that our method is highly effective in improving learning efficiency in environments with fewer parameters and lower quantization bits.

%

\begin{figure}[t]
    \centering
    \begin{subfigure}[b]{0.23\textwidth}
        \centering
        \includegraphics[width=\textwidth]{imgs/baseline-train-2d.png}
        \caption{}
        \label{landscape_baseline_b}
    \end{subfigure}
    \begin{subfigure}[b]{0.23\textwidth}   
        \centering 
        \includegraphics[width=\textwidth]{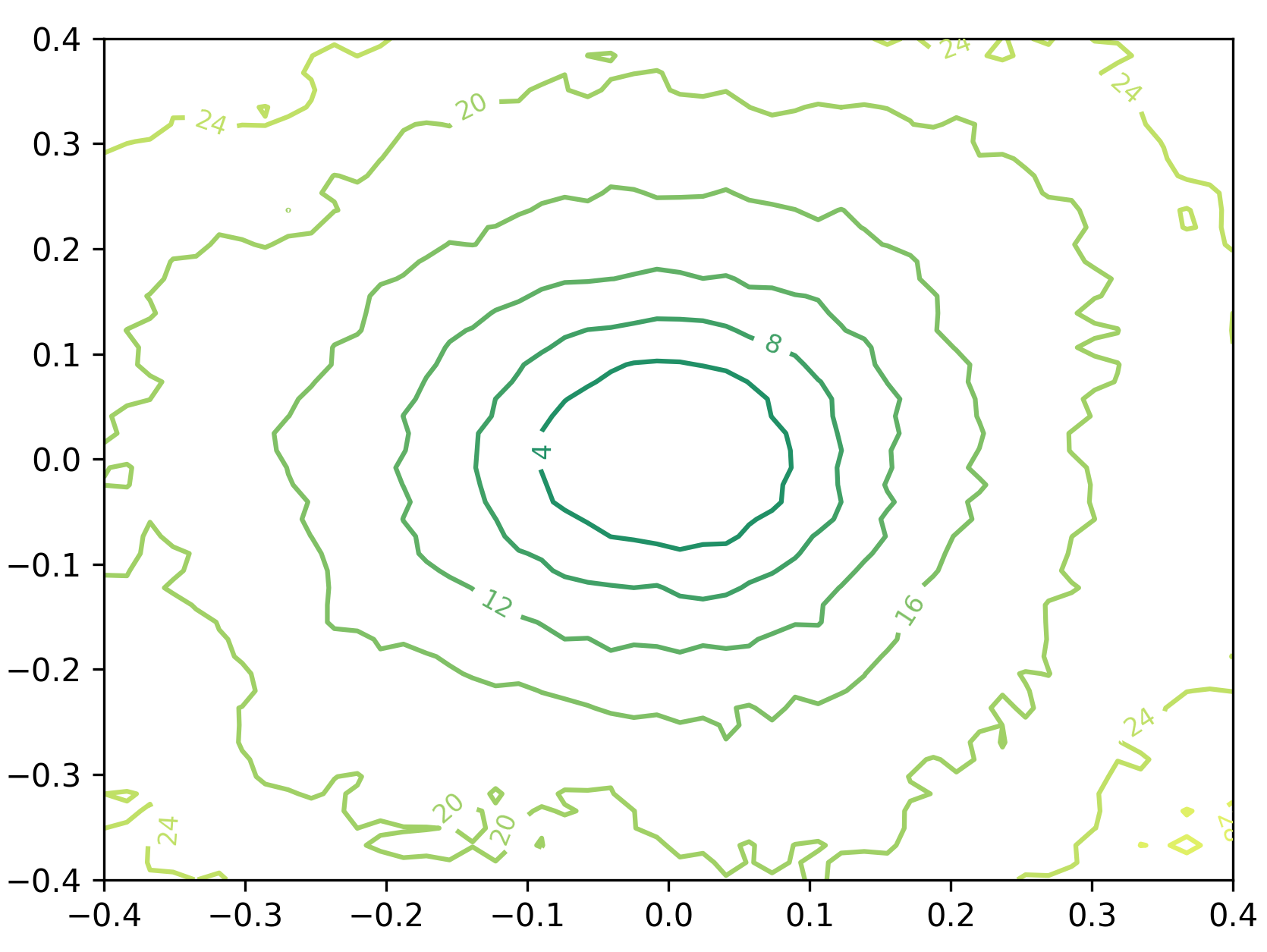} 
        \caption{}
        \label{fig:landscape-dual}
    \end{subfigure}
    \begin{subfigure}[b]{0.23\textwidth}  
        \centering 
        \includegraphics[width=\textwidth]{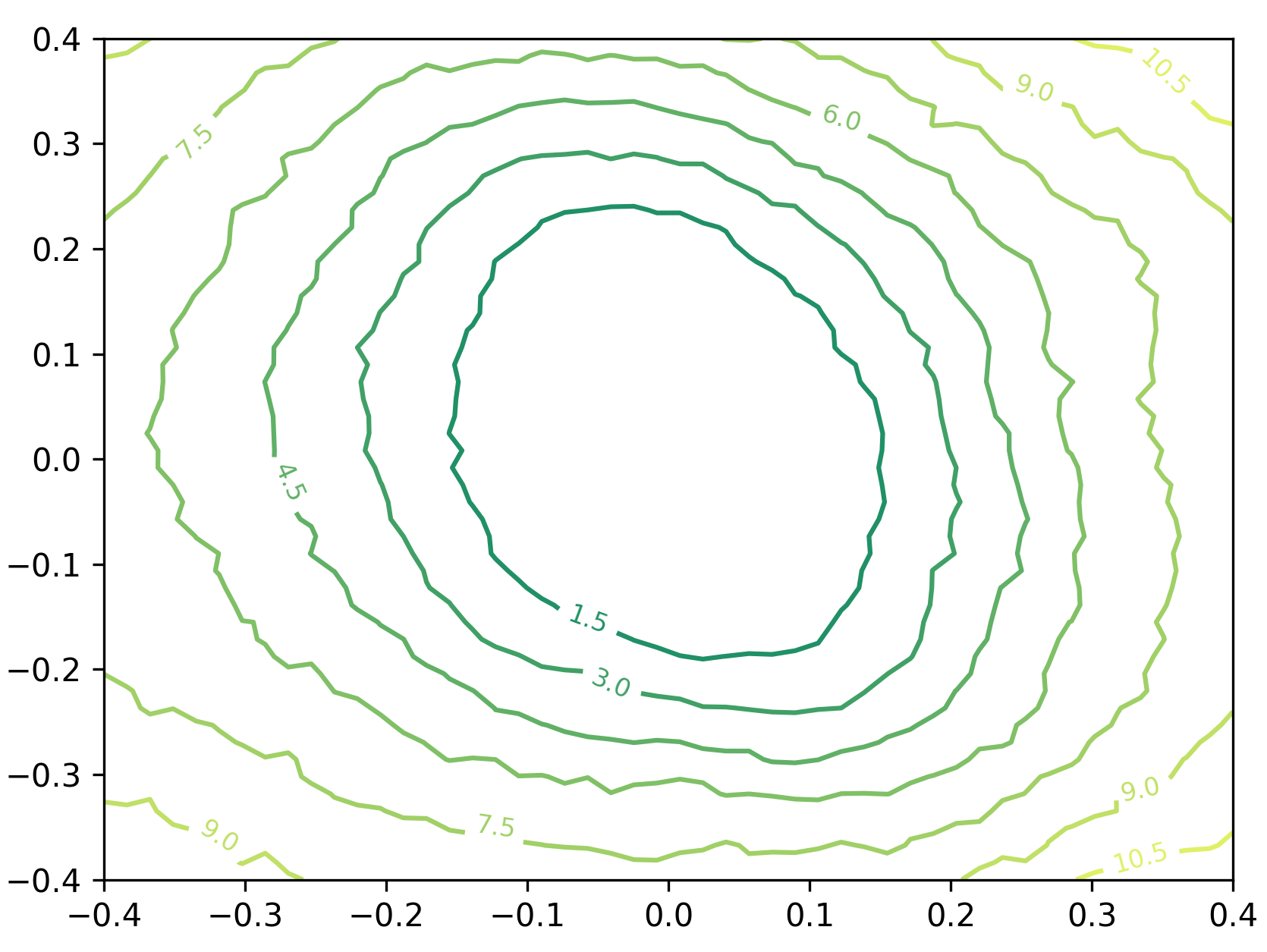} 
        \caption{}
        \label{fig:landscape-res}
    \end{subfigure}
    \begin{subfigure}[b]{0.23\textwidth}   
        \centering 
        \includegraphics[width=\textwidth]{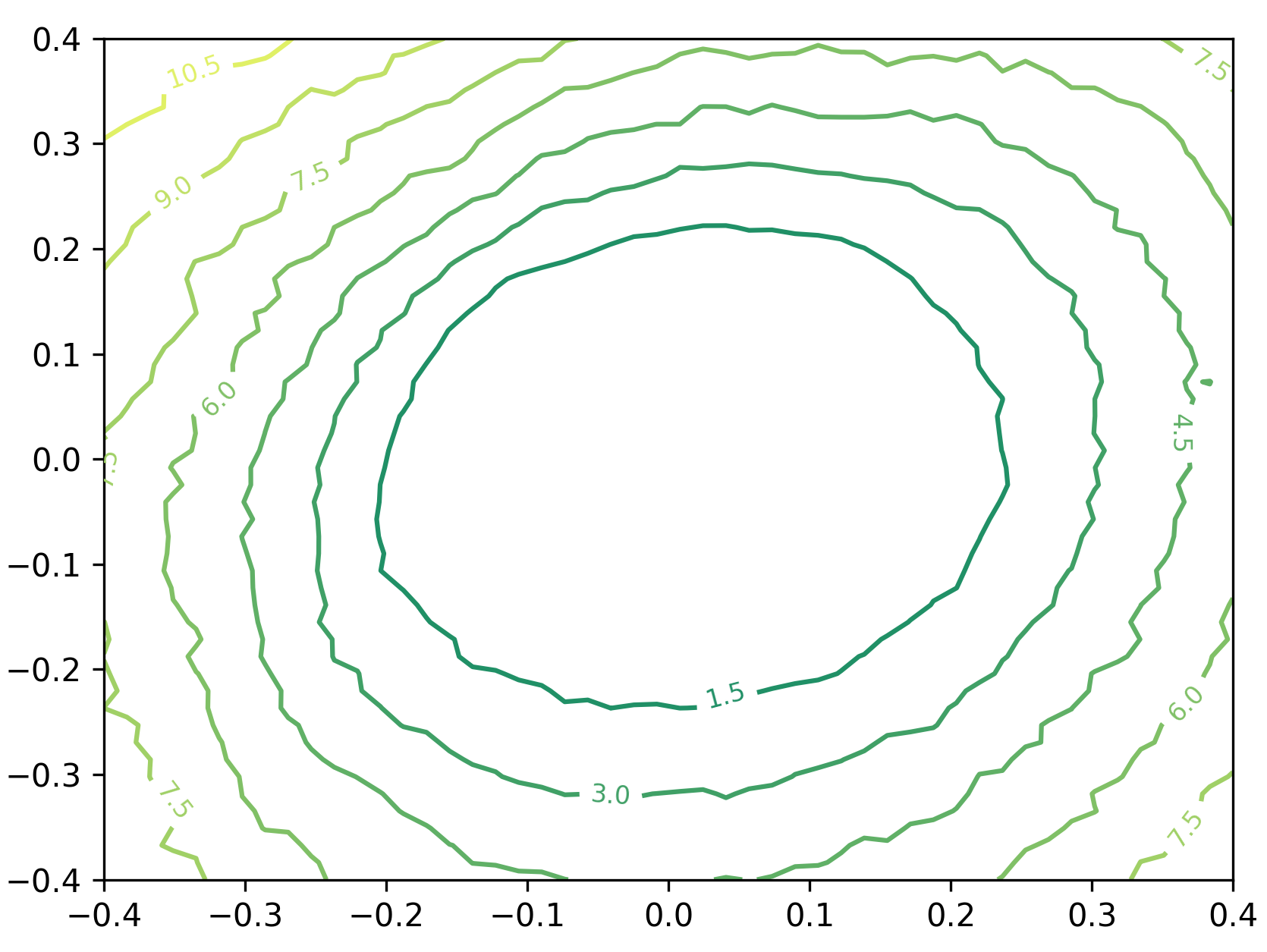} 
        \caption{}
        \label{fig:bd-net}
    \end{subfigure}
    
    \caption{Loss landscapes on the CIFAR-100 dataset: (a) 
    Binarized MobileNet V1 (same as~\Cref{landscape_baseline}), (b) Binarized MobileNet V1 with 1.58-bit conv, (c) Binarized MobileNet V1 with pre-BN residual, and (d) Our proposed method.}
    \label{fig:loss_landscape_2d}
    \vspace{-0.3cm}
\end{figure}

\vspace{0.5em}
\noindent\textbf{Loss Landscape.}

To assess the effect of our proposed method on training stability, we visualize the loss landscape using a random filter-normalized direction in ~\Cref{fig:loss_landscape_2d}. 

Comparing with the binarized MobileNet V1 shown in \Cref{landscape_baseline_b}, the landscape with 1.58-bit conv~\Cref{fig:landscape-dual} appears significantly smoother. 
This demonstrates how the representational ability of a binarized model impacts training. Our 1.58-bit conv approach enhances the model's representational ability in a parameter-efficient manner, and increases training stability. 

\Cref{fig:landscape-res} demonstrates how effectively the loss landscape can be smoothed by the pre-BN residual which effectively reduces the condition number. It slightly increases the computational cost without model parameter increase, but it has a significant effect on smoothing the loss landscape.
It is an effective structural compensation that mitigates the challenge originated from the structural limitations of DW convs. 
BD-Net, which incorporates both 1.58-bit conv and pre-BN residuals, demonstrates a much smoother loss landscape, enhancing both training stability and generalization capabilities. The landscape of BD-Net shows a level of smoothness very similar to that of full-precision depicted in \Cref{landscape_32}.

We further analyze the 3D loss landscapes of the baseline and BD-Net, revealing that BD-Net exhibits a smoother and more stable surface during optimization (see Appendix B).

\section{Conclusion}
We addressed the fundamental challenge of integrating binary DW convs into BNNs through comprehensive analysis. Our BD-Net introduces pre-BN residual connections and 1.58-bit convs, enabling the first successful binarization of DW convs in BNNs. Our method achieves the lowest computational cost with only 33M OPs on ImageNet while delivering up to 9.3 p.p. accuracy improvements across multiple datasets on MobileNet V1, demonstrating the potential for truly efficient binary neural networks. Furthermore, we extended our approach to ShuffleNet V1 and MobileNet V3, demonstrating its adaptability to other network architectures.

\section{Acknowledgments}
This work was supported by Institute of Information \& communications Technology Planning \& Evaluation (IITP) grant funded by the Korea government (MSIT)(RS-2019-II190421, Artificial Intelligence Graduate School Program (Sungkyunkwan University), 20\%),  (No.1711195788, Development of Flexible SW/HW Conjunctive Solution for on-edge self-supervised learning, 20\%), (No.RS-2025-25442569, AI Star Fellowship Support Program (Sungkyunkwan Univ.), 20\%), National Research Foundation of Korea (NRF) grant funded by the Korea government (MEST)(RS-2024-00352717, 20\%) and  IITP (Institute of Information \& Communications Technology Planning \& Evaluation)-ITRC (Information Technology Research Center) grant funded by the Korea government (Ministry of Science and ICT)(IITP-2026-RS-2024-00437633, 20\%).

\bibliography{aaai2026}

\clearpage
\setcounter{page}{1}
\onecolumn
\appendix

\begin{center}
    \large \textbf{Supplementary Material for}
    
    \vspace{0.05cm}
    
    \LARGE \textbf{BD-Net: Has depth-wise convolution ever been applied in \\ Binary Neural Networks?}
\end{center}
\vspace{0.7cm}

\section{Appendix A: Implementation Details}
\label{appendix:A}
\subsection{CIFAR-10 and CIFAR-100}
\label{sec:rationale}
The CIFAR-10 dataset consists of 50,000 training and 10,000 testing images of size 32$\times$32 across 10 classes. CIFAR-100 has the same image size and number of samples but contains 100 classes.
In MobileNet V1, 3$\times$3 regular convolutions from the AdaBin architecture are replaced with 3$\times$3 depth-wise convolutions. The initial learning rate is set to 0.1 and decayed using Cosine Annealing. The model is trained with a batch size of 256 over 400 epochs using the SGD optimizer with a momentum of 0.9. During training, 4 pixels of padding are added to each side, and 32$\times$32 random crops, along with random horizontal flips, are applied. The results reported in the tables are the average of two runs conducted with different random seeds.

\subsection{STL-10, Tiny ImageNet, and Oxford Flowers-102 Datasets}
The STL-10 dataset consists of 5,000 training and 8,000 testing images with a resolution of 96$\times$96 pixels, categorized into 10 classes. 
The Tiny ImageNet dataset contains 100,000 training and 10,000 testing images with a resolution of 64$\times$64 pixels, categorized into 200 classes. 
The Flowers-102 dataset consists of 1,020 training and 6,149 testing images of varying resolutions, categorized into 102 classes.  

STL-10 and Tiny ImageNet follow the same experimental settings as CIFAR-10 and CIFAR-100. For Oxford Flowers-102, with an input size of $224 \times 224$, we set the stride of the first layer to 2 and replace the $1 \times 1$ point-wise convolutions with dual convolutions to accommodate the higher resolution. The results reported in the tables are the average of two runs conducted with different random seeds.

\subsection{ImageNet}
ImageNet is a large-scale benchmark dataset, which consists of  over 1.28 million training images, 50,000 test images, categorized into 1,000 classes. 
The baseline is based on ReActNet, but 3$\times$3 regular convolutions are substituted with 3$\times$3 depth-wise convolutions.
In BD-Net, 1.58-bit convolutions are utilized in binary 3$\times$3 depth-wise convolutions and binary 1$\times$1 convolutions.

For the training of BD-Net, a two-step training strategy is adopted. 
In the first step, the network is trained using binary activations and real-valued weights.
In the subsequent step, the trained weights are used as initial values, and both weights and activations are binarized for fine-tuning. 
Throughout all training phases, the Adam optimizer and a linear learning rate decay scheduler are used, starting with an initial rate of 5e-4.
In each step, both the number of epochs and the batch size are set to 256. Weight decay is set to 1e-5 in the initial step and reduced to zero in the second step.
The distributional loss is utilized as the objective for optimization in both steps, replacing the traditional cross-entropy loss.
The results reported in the tables are the average of two runs conducted with different random seeds.

\section{Appendix B: Analysis}
\label{appendix:B}
\subsection{3D landscape loss}
\begin{figure}[H]
    \centering
    \begin{subfigure}[t]{0.3\textwidth} 
        \centering
        \includegraphics[width=\textwidth,height=3.5cm]{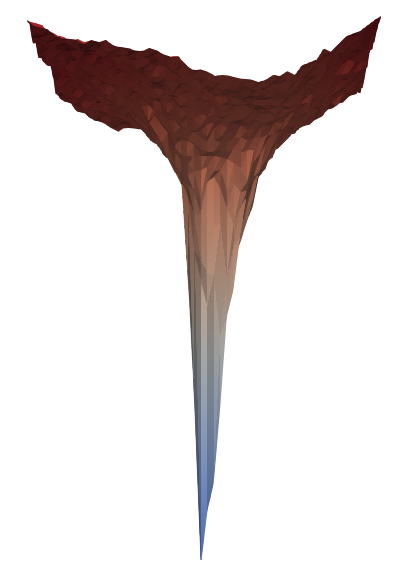} 
        \caption{Binarized MobileNet}
        \label{fig:baseline_3d}
    \end{subfigure}
    \hspace{0.5cm}
    \begin{subfigure}[t]{0.3\textwidth} 
        \centering
        \includegraphics[width=\textwidth, height=3.5cm]{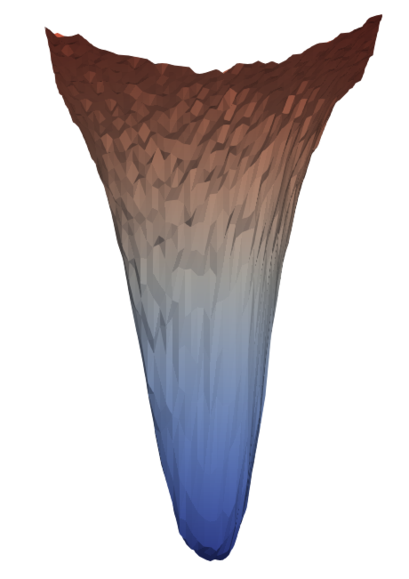} 
        \caption{Ours}
        \label{fig:ours_3d}
    \end{subfigure}
    \caption{3D Landscape}
    \label{fig:3d_landscape}
\end{figure}

We present the 3D loss landscapes of both the baseline and BD-Net in Figure~\ref{fig:3d_landscape}. As shown, the baseline (Figure~\ref{fig:baseline_3d}) exhibits a sharp curvature near local minima, indicating susceptibility to instability from slight weight perturbations. In contrast, BD-Net (Figure~\ref{fig:ours_3d}) demonstrates a smoother curvature, highlighting its improved robustness and stability during training.

\subsection{Ablation Study}
To evaluate the effectiveness of the Pre-BN residual and multiple depth-wise convolutions in BD-Net, we conduct an ablation study by increasing the number of depth-wise convolutions under both the Baseline and Pre-BN settings. In this experiment, the Baseline refers to the MobileNet V1 AdaBin architecture in which the 3$\times$3 regular convolutions are replaced with 3$\times$3 depth-wise convolutions. The Pre-BN setting extends the Baseline by adding the proposed Pre-BN residual to each depth-wise convolution layer. The results are summarized in Table~\ref{tab:ablation_combined}. Here, Single, Dual, Triple, and Quad denote configurations with 1, 2, 3, and 4 depth-wise convolutions applied within each depth-wise convolution layer, respectively.

In the Baseline, increasing the number of depth-wise convolutions improves accuracy. However, the performance gain becomes marginal beyond Triple, relative to the additional computational cost. Similarly, in the Pre-BN setting, accuracy consistently improves as the number of depth-wise convolutions increases. These results indicate that the Pre-BN residual and multiple depth-wise convolutions complement each other, leading to greater performance gains when combined.

\begin{table}[H]
\centering
\large
\scalebox{1.1}{  
\begin{tabular}{lcccc}
\toprule
\textbf{Method} & \textbf{Single} & \textbf{Dual} & \textbf{Triple} & \textbf{Quad} \\
\midrule
\textbf{Baseline} & 54.94 & 56.18 & 56.77 & 57.46 \\
\textbf{pre-BN} & 56.93 & 58.28 & 58.91 & 59.26 \\
\bottomrule
\end{tabular}
}
\caption{Ablation study on MobileNet V1 with CIFAR-100. Comparison of the baseline and pre-BN models across different numbers of depth-wise convolutions (Single, Dual, Triple, Quad). Results are reported in accuracy (\%).}
\label{tab:ablation_combined}
\vspace{-2mm}
\end{table}

\vspace{0.7em}
\noindent
\subsection{Training and Validation Accuracy Curve}
\begin{figure}[H]
    \centering
    \includegraphics[width=0.7\textwidth]{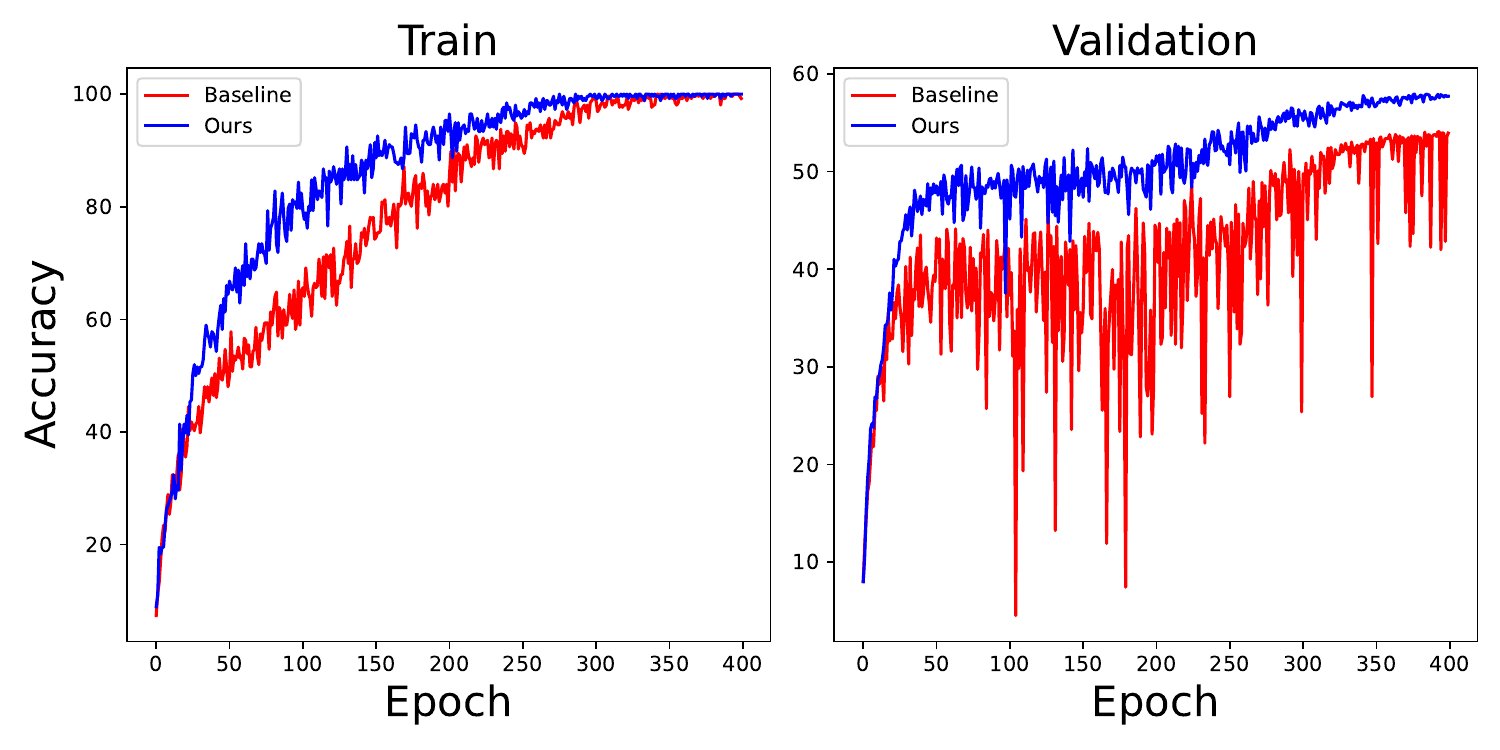} 
    \caption{Comparison of training and validation accuracy curves between the baseline and our model on CIFAR-100.}
    \label{fig:cifar_accuracy_plot}
\end{figure}

\Cref{fig:cifar_accuracy_plot} shows the training and validation accuracy curves. For training accuracy, both our method and the baseline achieve comparable final accuracy, but our method converges faster and more efficiently.  
In contrast, a clear difference appears in validation accuracy. The baseline exhibits significant fluctuations in validation accuracy despite steadily increasing training accuracy, indicating overfitting and poor generalization. Although the validation data differs only slightly from the training data, the baseline is highly sensitive to these minor shifts. This suggests that the rough loss landscape of the baseline hinders its ability to find weights that generalize well.  

In contrast, our proposed model achieves stable improvements in both training and validation accuracy without notable fluctuations. These results show that our method not only accelerates convergence but also enhances generalization by effectively smoothing the loss landscape.

\clearpage

\section{Appendix C: Hardware Performance Evaluation}

To evaluate the practical latency of our proposed BD-Net, we conducted experiments on a real mobile device (Galaxy S21+), using a single CPU thread with the Larq Compute Engine (LCE). While LCE natively supports binary regular convolution operations, it currently does not support binary depth-wise convolutions. As a result, we implemented our BD-Net using 32-bit depth-wise convolutions instead.

The latency of binary depth-wise convolution was conservatively estimated as one-third of its 32-bit counterpart. This estimate was derived based on our own empirical measurements: full-precision 3$\times$3 regular convolution took 21.03 ms, whereas binary 3$\times$3 regular convolution took only 2.60 ms. This nearly 8× speed-up reflects the substantial efficiency of binary operations. Since depth-wise convolutions require significantly fewer operations than regular convolution, we conservatively assumed a 3× speed-up for the binary depth-wise convolution over its 32-bit version.

Table~\ref{tab:operation_latency} summarizes the latency of various primitive operations. Despite the presence of the pre-BN residual, the total latency of dual binary depth-wise convolution combined with the residual add operation remains under one-third of the latency of a binary 3$\times$3 regular convolution. This highlights the hardware execution efficiency of our proposed structure.

\begin{table}[H]
\centering
\large
\scalebox{1.1}{
\begin{tabular}{lc}
\toprule
\textbf{Operations} & \textbf{Latency (ms)} \\
\midrule
32-bit 3$\times$3 Regular Conv & 21.03 \\
Binary 3$\times$3 Regular Conv & 2.60 \\
Binary 1$\times$1 Regular Conv & 0.38 \\
32-bit 3$\times$3 DW Conv & 0.96 \\
Binary 3$\times$3 DW Conv* & 0.32 \\
Pre-BN residual (ADD) & 0.17 \\
\bottomrule
\end{tabular}
}
\caption{Operation latency comparison at 56$\times$56 resolution with 128 input/output channels. Binary 3$\times$3 DW Conv values (*) are conservatively estimated at one-third the latency of 32-bit DW Conv, based on observed speed-up of binary regular conv.}
\label{tab:operation_latency}
\end{table}

Table~\ref{tab:total_latency} compares the end-to-end latency of several models. Notably, our implementation of \textbf{BD-Net-B$^\dagger$}, which replaces binary depth-wise convolutions with 32-bit depth-wise convolutions, already outperforms ReActNet in terms of latency. The latency of 32-bit depth-wise convolutions in BD-Net-B$^\dagger$ is 5.81 ms, and the latency of other operations is 9.13 ms. Based on the estimated latency of binary depth-wise convolutions (1.94 ms), the total latency of the original model BD-Net-B$^*$ is projected to be 11.07 ms. This is approximately 60\% lower than ReActNet under the same hardware condition. If binary depth-wise convolution were fully supported by the backend, the total runtime latency could be further reduced beyond current estimates.

\begin{table}[H]
\centering
\large
\scalebox{1.1}{
\begin{tabular}{lcc}
\toprule
Backbone & \textbf{Method} & \textbf{Latency (ms)} \\
\midrule
\multirow{2}{*}{ResNet 18} & Full-Precision & 91.18 \\
& ReActNet & 16.76 \\
\midrule
\multirow{4}{*}{MobileNet V1} & Full-Precision & 29.28 \\
& ReActNet & 27.66 \\
& BD-Net-B$^\dagger$ & 14.94 \\
& \textbf{BD-Net-B$^*$} & \textbf{11.07} \\
\bottomrule
\end{tabular}
}
\caption{Latency comparison (in milliseconds). 
\textbf{BD-Net-B$^\dagger$} uses 32-bit DW Convs instead of binary DW Convs, while \textbf{BD-Net-B$^*$} denotes the latency estimate of our original model with binary DW Convs.}
\label{tab:total_latency}
\end{table}

While ReActNet demonstrates latency improvements on ResNet-based backbones, its advantage diminishes on MobileNet V1 due to its use of binary 3$\times$3 regular convolutions in place of the original depth-wise convolutions. As shown in Table~\ref{tab:operation_latency}, binary regular convolutions are substantially slower than depth-wise convolutions, even when using full-precision.

In contrast, BD-Net preserves the efficient depth-wise convolutional structure and successfully binarizes it. It combines dual binary depth-wise convolutions with a pre-BN residual connection to improve expressiveness without incurring high latency. The estimated latency of BD-Net-B$^*$ suggests that our architecture is better suited for edge deployment scenarios requiring low-latency inference.

Due to the channel-wise independence of depth-wise convolutions, binary depth-wise convolution operations are well suited for execution within the CPU cache. At typical spatial resolutions, the bit-packed input, weights, and outputs for each channel can often fit entirely within cache memory, enabling cache-resident execution.

Additionally, the same channel-wise property enables the structural fusion of two binary depth-wise convolutions and a residual addition. Each operation—convolution, convolution, and addition—can be executed per channel without inter-channel dependencies. While our current implementation does not support this fusion, it is feasible to implement at the kernel level. Doing so would eliminate intermediate memory writes between layers and further reduce memory latency. Therefore, the actual runtime latency of BD-Net could be further improved in future backend implementations that support this fused execution model.

\section{Appendix D: Visualization of 1.58-bit}
In this section, we provide additional visualizations of the validation images on ImageNet. The 1.58-bit representation demonstrates better representational power than Otsu's algorithm.

\begin{figure*}[h]
    \centering
    \includegraphics[width=0.9\textwidth]{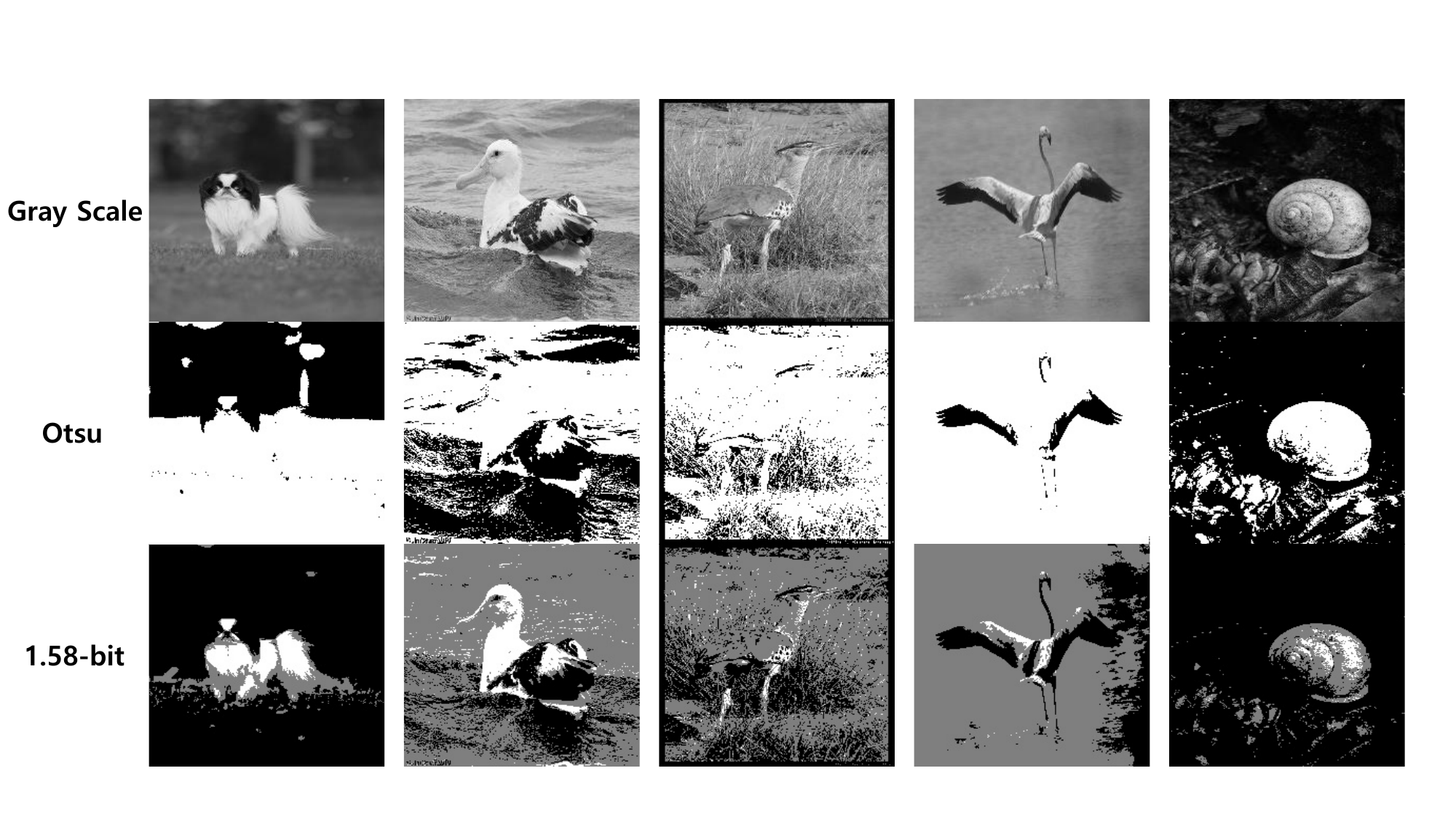}
    \caption{Image visualizations on the ImageNet validation dataset are presented as follows: the first row shows images converted to grayscale, the second row applies Otsu's algorithm, and the last row illustrates the 1.58-bit representation.}
    \label{Visualization of the validation images on ImageNet}
\end{figure*}

\end{document}